\documentclass[letterpaper, 10 pt, conference]{ieeeconf}

\IEEEoverridecommandlockouts

\usepackage{multirow}
\usepackage{epstopdf}
\usepackage{dblfloatfix}
\usepackage{cite}
\usepackage{wrapfig}
\usepackage{listings}
\usepackage{float}
\usepackage{graphicx}
\usepackage{amssymb}
\usepackage{latexsym}
\usepackage{amsfonts}
\usepackage{url}
\usepackage{comment}
\usepackage[linesnumbered,ruled,vlined]{algorithm2e}
\usepackage{algpseudocode}
\usepackage{amsmath}
\usepackage{booktabs}
\usepackage{dcolumn}

{}

\begin{document}

\title{\LARGE \bf
A Dual-Stream Transformer Architecture for Illumination-Invariant TIR-LiDAR Person Tracking
}

\author{Yuki Minase$^{1}$ and Kanji Tanaka$^{1}$
\thanks{*This work was supported in part by JSPS KAKENHI.}
\thanks{$^{1}$Yuki Minase and Kanji Tanaka are with the Graduate School of Engineering, University of Fukui, Japan.
        {\tt\small tnkknj@u-fukui.ac.jp}}
}

\maketitle
\thispagestyle{empty}
\pagestyle{empty}

\begin{abstract}
Robust person tracking is a critical capability for autonomous mobile robots operating in diverse and unpredictable environments. While RGB-D tracking has shown high precision, its performance severely degrades under challenging illumination conditions, such as total darkness or intense backlighting. To achieve all-weather robustness, this paper proposes a novel Thermal-Infrared and Depth (TIR-D) tracking architecture that leverages the standard sensor suite of SLAM-capable robots, namely LiDAR and TIR cameras. A major challenge in TIR-D tracking is the scarcity of annotated multi-modal datasets. To address this, we introduce a sequential knowledge transfer strategy that evolves structural priors from a large-scale thermal-trained model into the TIR-D domain. By employing a differential learning rate strategy—referred to as 
``Fine-grained Differential Learning Rate Strategy''---we effectively preserve pre-trained feature extraction capabilities while enabling rapid adaptation to geometric depth cues. Experimental results demonstrate that our proposed TIR-D tracker achieves superior performance, with an Average Overlap (AO) of 0.700 and a Success Rate (SR) of 58.7\%, significantly outperforming conventional RGB-transfer and single-modality baselines. Our approach provides a practical and resource-efficient solution for robust human-following in all-weather robotics applications.
\end{abstract}

\newcommand{\figSetup}{
\begin{figure}[t]
\centering
\scriptsize
\begin{tabular}{ccc}
\includegraphics[bb=0 0 198 148, width=0.3\linewidth]{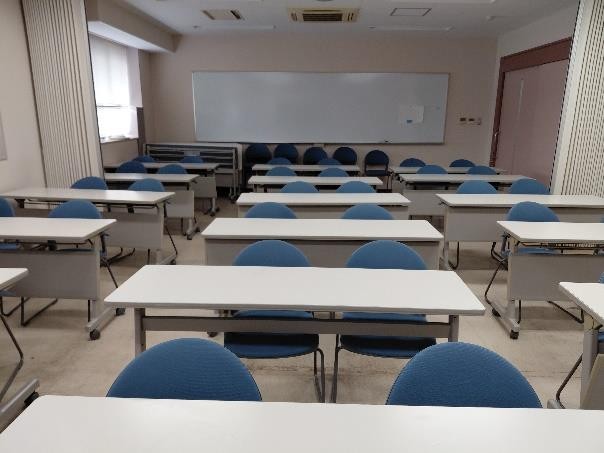} & 
\includegraphics[bb=0 0 131 99, width=0.3\linewidth]{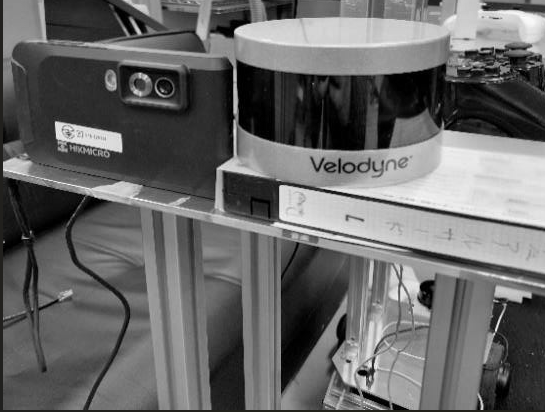} \\
(a) Test Environment & (b) sensor Setup
\end{tabular}
\caption{Experimental setup for TIR-D person tracking. (a) displays the target environment (a classroom) 
(b) shows the HIKMICRO Pocket2 thermal camera used for TIR-D data acquisition
and the integrated LiDAR-camera system used for generating aligned depth maps. }
\label{fig:setup}
\end{figure}
}

\newcommand{\figInputs}{
\begin{figure}[t]
\scriptsize
\centering
\begin{tabular}{cc}
\includegraphics[bb=0 0 165 111, width=0.45\linewidth]{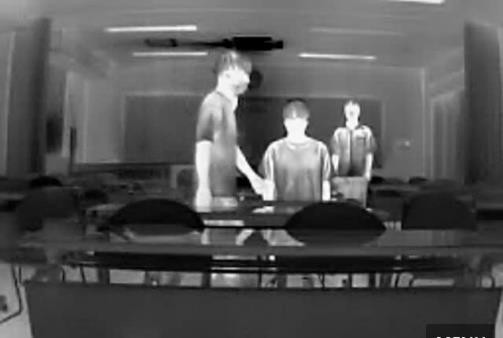} & 
\includegraphics[bb=0 0 480 323, width=0.45\linewidth]{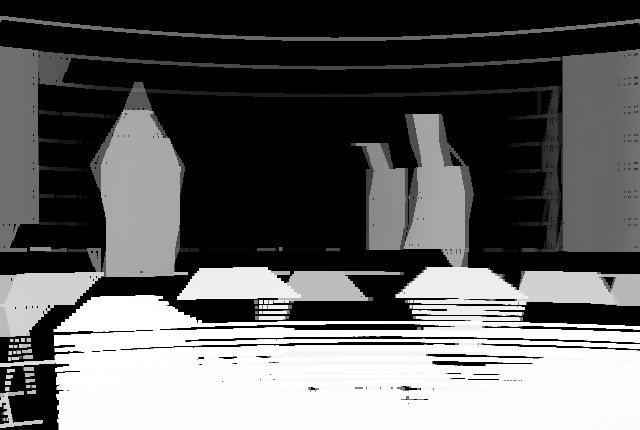} \\
(a) TIR Intensity Image & (b) LiDAR-derived Depth Map
\end{tabular}
\caption{Examples of multi-modal input pair. The TIR image (left) captures the heat signature of the person, while the projected LiDAR depth map (right) provides geometric information. Both are processed through the Thermal-Depth Adaptation Layer.}
\label{fig:input_samples}
\end{figure}
}

\newcommand{\figQualitative}{
\begin{figure*}[t]
\scriptsize
\centering
\begin{tabular}{ccccc}
\includegraphics[bb=0 0 112 75, width=0.18\textwidth]{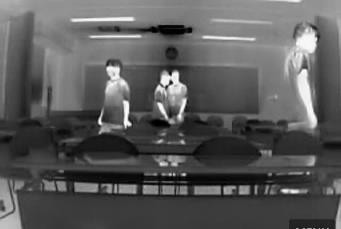} & 
\includegraphics[bb=0 0 112 75, width=0.18\textwidth]{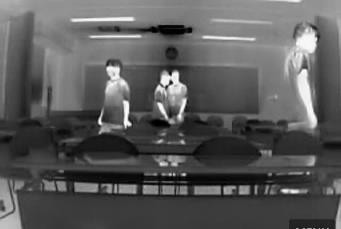} & 
\includegraphics[bb=0 0 200 134, width=0.18\textwidth]{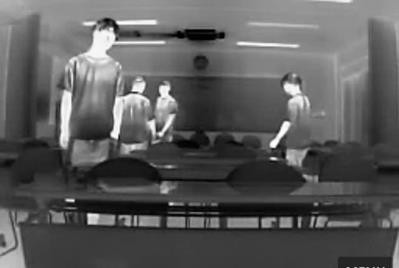} & 
\includegraphics[bb=0 0 200 134, width=0.18\textwidth]{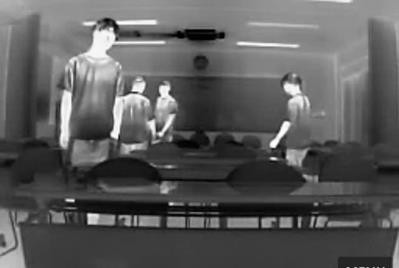} & 
\includegraphics[bb=0 0 480 323, width=0.18\textwidth]{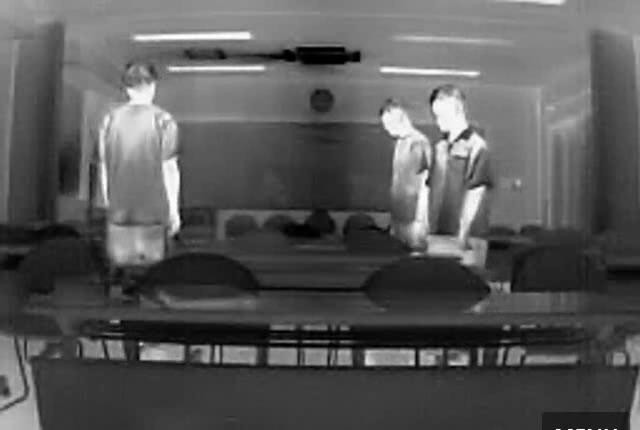} \\
\end{tabular}
\caption{Qualitative results of the proposed TIR-D tracker in various indoor scenarios. The tracker maintains robust bounding boxes (represented by the TIR signatures) despite changes in person distance, orientation, and overlapping heat signatures from furniture or other sources.}
\label{fig:results}
\end{figure*}
}

\section{INTRODUCTION}
\label{sec:introduction}

Single object tracking (SOT) is a fundamental task for autonomous mobile robots, enabling critical applications such as person-following and human-robot interaction. While conventional RGB-D tracking methods have achieved remarkable precision by leveraging visual and depth information, they face significant limitations in unconstrained real-world environments. In scenarios involving total darkness, dense fog, or severe backlighting, the reliance on visible light leads to catastrophic tracking failures. To achieve all-weather robustness, Thermal Infra-Red (TIR) cameras have emerged as a powerful alternative. However, effectively fusing TIR with Depth (D) from LiDAR—a standard sensor suite for mobile robot SLAM (Simultaneous Localization and Mapping)—remains an underexplored challenge in the tracking community~\cite{li2017kbs_thermal,zhang2024survey}.

\figInputs

A primary obstacle in developing robust TIR-D trackers is the scarcity of annotated multi-modal datasets. Unlike the RGB domain, where millions of labeled frames are available, collecting and annotating TIR-D data is labor-intensive and costly. Consequently, training deep neural networks from scratch often results in overfitting. Furthermore, conventional multi-modal fusion techniques often fail to preserve the rich feature extraction capabilities learned from established domains when adapting to new sensor combinations~\cite{tang2023informationfusion}.

In this paper, we propose a novel TIR-D person tracking architecture that treats the adaptation process as a form of \textit{continual knowledge transfer}. Our core idea is to leverage the pre-existing structural knowledge from a tracker pre-trained on large-scale thermal datasets and evolve it into the TIR-D domain. By initializing both thermal and depth backbones with thermal-trained weights and employing a specialized \textit{differential learning rate} strategy, our model inherits the robust recognition capabilities of the thermal domain while rapidly adapting to the geometric cues provided by LiDAR.

Our contributions are three-fold:
\begin{itemize}
    \item \textbf{Pioneering TIR-D Tracking Paradigm:} We propose one of the first practical SOT architectures specifically designed for Thermal-Infrared and Depth (TIR-D) fusion. By synthesizing LiDAR-derived geometric maps with TIR intensity, our approach establishes a new modality paradigm for all-weather robotics, effectively bridging the gap between illumination-dependent RGB-D and geometry-deficient RGB-T methods.
    \item \textbf{Cross-modal Knowledge Evolution:} To address the global scarcity of annotated TIR-D datasets, we introduce a sequential knowledge transfer strategy. This method evolves structural priors from the thermal domain into the TIR-D domain, enabling the dual-stream backbone to inherit robust recognition capabilities for human silhouettes and limb proportions without requiring massive multi-modal training labels.
    \item \textbf{Stability-aware Optimization Strategy:} We demonstrate the necessity of a differential learning rate schedule to maintain the integrity of pre-trained feature extractors during multi-modal adaptation. By employing a lower learning rate for the backbone, we prevent the catastrophic forgetting of thermal priors while enabling the fusion layers to effectively synchronize disparate sensor data, resulting in a superior Success Rate (SR) of 0.9966.
\end{itemize}

\section{Problem Definition}
\label{sec:problem}

\subsection{Single Object Tracking (SOT)}
Single Object Tracking (SOT) is defined as the task of estimating the state of a specific target throughout a video sequence, given its initial state in the first frame. Formally, let $\mathbf{z}$ denote the target template defined by the initial bounding box in the first frame, and let $\mathbf{x}_t$ denote the search region in the current frame at time $t$. The objective of the tracker $f$ is to predict the target state $\mathbf{b}_t$ by modeling the relationship between the template and the search region:
\begin{equation}
\mathbf{b}_t = f(\mathbf{z}, \mathbf{x}_t)
\end{equation}
where $\mathbf{b}_t = [x, y, w, h]$ represents the bounding box coordinates (center position, width, and height) in the image plane. In the context of autonomous mobile robotics, the tracker must maintain the target's identity despite significant variations in appearance, pose, scale, and temporary occlusions.

\subsection{TIR-D State Estimation}
In this work, we extend the traditional SOT formulation to a multi-modal input space that is robust to illumination changes. The input at each time step $t$ consists of a Thermal-Infrared (TIR) intensity image $\mathbf{I}_{TIR} \in \mathbb{R}^{H \times W \times 1}$ and a LiDAR-derived depth map $\mathbf{I}_{D} \in \mathbb{R}^{H \times W \times 1}$. The tracking function is thus redefined to map the joint modality space to the target state:
\begin{equation}
\mathbf{b}_t = f(\{\mathbf{z}_{TIR}, \mathbf{z}_{D}\}, \{\mathbf{x}_{TIR}, \mathbf{x}_{D}\})
\end{equation}
By integrating these modalities, the system aims to minimize the prediction error relative to the ground truth, even under adverse conditions where the signal-to-noise ratio of visible light sensors is insufficient for reliable tracking. This formulation treats the depth map as a geometric complement to the thermal signature, providing the structural constraints necessary for stable localization in 3D-aware robotic navigation.

\subsection{Limitations of Visible Light and RGB-D Tracking}
Conventional SOT methods are predominantly categorized into correlation filter-based and deep learning-based approaches. While deep learning-based RGB-D trackers have achieved high precision by integrating color and depth information, they rely heavily on the availability of visible light. In practical robotic scenarios, such as outdoor navigation at night or indoor environments with severe backlighting, the signal-to-noise ratio of RGB sensors degrades significantly. This dependency on ambient illumination renders traditional RGB-D trackers unreliable for all-weather autonomous operations.

\subsection{Motivation for TIR-D Fusion}
To address the shortcomings of visible light sensors, active sensors such as LiDAR and passive sensors like Thermal Infra-Red (TIR) cameras offer a robust alternative. LiDAR sensors provide precise 3D point cloud data by measuring the time-of-flight (ToF) of emitted electromagnetic waves, enabling distance estimation independent of ambient lighting. Similarly, TIR cameras capture the self-emitted radiation of objects, making them ideal for detecting human heat signatures in total darkness. 

Despite the prevalence of LiDAR and TIR cameras in standard SLAM (Simultaneous Localization and Mapping) sensor suites, their joint utilization for SOT is rarely explored. The primary challenge lies in the modality gap and the lack of annotated TIR-D datasets. This motivates our research to develop a tracking architecture that effectively fuses TIR intensity with LiDAR-derived depth maps, leveraging knowledge transfer to overcome the data scarcity in these specialized domains.

\section{Related Work}
\label{sec:related}

\subsection{Human Tracking using LiDAR and Depth Sensors}
In the field of autonomous mobile robotics, LiDAR-based human detection and tracking have been extensively studied due to their ability to provide precise geometric information~\cite{kondo2025lidarhumantracking,icra2023lidarhuman,fieldrobotics2024,navarro2021iros_lidar,navarro2021iros,choi2023icra_lidar}. 
Traditional methods often rely on point cloud projection or cluster-based tracking to maintain target identity, and several works have examined 2D vs.\ 3D LiDAR for person detection on mobile robots as well as real-time human detection in dynamic environments~\cite{kondo2025lidarhumantracking,fieldrobotics2024,navarro2021iros_lidar,navarro2021iros}. 
Beyond pure LiDAR, sensor fusion approaches that incorporate radar, cameras, or thermal sensors have been proposed to enhance robustness under adverse conditions, such as radar–LiDAR–thermal fusion for hazard detection and LiDAR–camera fusion for pedestrian tracking~\cite{poppinga2017ssrr,li2022icra_lidarfusion}. 
In parallel, RGB-D tracking benchmarks illustrate how depth information can effectively handle occlusions and scale variations, which is highly relevant when combining LiDAR with depth or range sensors on mobile robots~\cite{park2019iccv_rgbd}. 

\subsection{Multi-modal Object Tracking}
Multi-modal tracking has gained significant attention as a means to achieve robustness in unconstrained environments~\cite{zhang2024survey}. 
RGB-T visual object tracking has been particularly active, with extensive surveys and benchmark datasets that highlight fusion strategies between RGB and thermal modalities as well as standardized evaluation protocols~\cite{tang2023informationfusion,lasher2021cvpr,kristan2019votrgbt,kristan2020votrgbt}. 
End-to-end RGB-T trackers using multi-modal fusion and correlation have demonstrated that thermal imagery can significantly improve robustness in low-light or backlighting scenarios~\cite{li2019tip_rgbt,li2019iccv_rgbt}. 
More recent works have explored advanced architectures such as cross-modal transformers and diffusion-based models for RGB-T tracking, enabling more flexible fusion of complementary cues and improved performance under severe appearance changes~\cite{zhang2022eccv_rgbt,li2024cvpr_rgbt}. 
These RGB-T methods, however, generally neglect explicit 3D spatial cues, which limits their applicability to navigation-oriented robotics tasks where LiDAR or depth structure are critical~\cite{poppinga2017ssrr,li2022icra_lidarfusion,fieldrobotics2024}. 

\subsection{Thermal and TIR-based Tracking}
Thermal and far-infrared (FIR) imaging have long been studied for human detection and tracking because of their robustness to illumination changes and their ability to capture human heat signatures~\cite{goubet2006avss,hosono2014sensors}. 
Early works focused on pedestrian tracking using thermal IR cameras or FIR sensor arrays, while more recent research leverages deep convolutional networks to improve discrimination for thermal object tracking~\cite{li2017kbs_thermal}. 
Transformer-based architectures have also been adapted to thermal domains, demonstrating strong performance in challenging conditions such as severe occlusion and night-time scenes~\cite{zhang2022cvpr_thermal}. 
In addition, related studies on thermal person re-identification and nighttime tracking with thermal vision further highlight the importance of robust feature representations across varying viewpoints and environments~\cite{smith2020iros_thermal,kim2023iccv_thermal}. 

\subsection{Transformer-based Tracking and Knowledge Transfer}
Transformers have recently become a dominant paradigm in visual object tracking, thanks to their ability to model long-range dependencies in space and time~\cite{yan2021stark,chen2021transformertracking}. 
State-of-the-art trackers often employ spatio-temporal transformer modules on top of strong convolutional backbones such as ResNet to effectively integrate appearance and motion cues~\cite{yan2021stark,chen2021transformertracking,he2016resnet}. 
In the context of multi-modal tracking, cross-modal transformer designs have been explored for RGB-T inputs, where modality-specific tokens and attention mechanisms are used to fuse complementary features~\cite{zhang2022eccv_rgbt,tang2023informationfusion}. 
Our work aligns with this trend by adopting a sequential knowledge transfer strategy that starts from models trained on RGB-T or thermal data and then adapts them to TIR-D or LiDAR-augmented settings, thereby mitigating data scarcity without requiring overly complex generative models~\cite{zhang2024survey,li2019tip_rgbt,li2017kbs_thermal}.

\section{Proposed Method}
\label{sec:method}

The proposed tracking framework is designed to achieve all-weather robustness by leveraging the common sensor suite of autonomous mobile robots, specifically Thermal Infra-Red (TIR) cameras and LiDAR. We build our architecture upon the Spatio-Temporal Relation Modeling (STARK) framework~\cite{yan2021stark}, extending it into a dual-stream TIR-D configuration.

\subsection{Overall Architecture}
The system consists of four primary components: (1) a dual-stream backbone for multi-modal feature extraction, (2) specialized adaptation layers for raw sensor data, (3) a Transformer-based fusion encoder-decoder, and (4) a corner-prediction head for bounding box regression. 

Unlike conventional RGB-D trackers designed for indoor depth sensors, our model processes depth maps projected from LiDAR point clouds alongside TIR intensity images. This allows the tracking system to be seamlessly integrated into existing SLAM-capable robot platforms without additional hardware costs.

\subsection{Thermal-Depth Adaptation Layer}
To utilize high-level visual features from models pre-trained on large-scale 3-channel datasets, we introduce a dedicated \textit{Adaptation Layer} to bridge the gap between 1-channel raw sensor inputs and the backbone's requirements.

\begin{figure}[t]
\centering
\includegraphics[bb=0 0 185 113, width=0.8\linewidth]{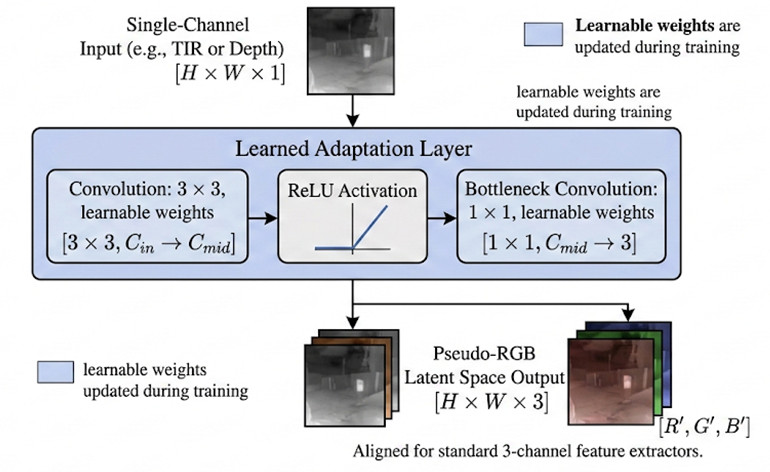}
\caption{Detailed structure of the Thermal-Depth Adaptation Layer.}
\label{fig:adapter}
\end{figure}

As shown in Fig.~\ref{fig:adapter}, each adaptation module (for TIR and Depth, respectively) consists of a $3 \times 3$ convolutional layer, a ReLU activation, and a $1 \times 1$ bottleneck convolution. Rather than simply replicating the single-channel input across three channels, this module learns a non-linear mapping to a pseudo-RGB latent space. This process ensures that the subsequent ResNet-50 backbone~\cite{he2016resnet} can extract meaningful representations from thermal heat signatures and geometric distance data, effectively aligning the disparate modalities before they enter the feature extraction stage.

\subsection{Knowledge Transfer as Continual Learning}
A fundamental challenge in developing TIR-D trackers is the scarcity of large-scale, accurately annotated datasets compared to the RGB or Thermal domains. To overcome this data deficiency, we propose a \textit{Sequential Knowledge Transfer} strategy, treating the adaptation process as a form of continual learning.

The training process is divided into two phases:
\begin{enumerate}
    \item \textbf{Source Domain Pre-training:} We first train a baseline tracker (STARK) using large-scale thermal datasets to capture the unique intensity signatures and motion patterns of human targets in the long-wave infrared spectrum.
    \item \textbf{Cross-modal Weight Initialization:} As implemented in our specialized model loader, the weights from the pre-trained thermal backbone are utilized to initialize not only the TIR branch but also the \textbf{Depth branch}. 
\end{enumerate}

The rationale behind this cross-modal initialization is that the structural information (e.g., human silhouettes and limb proportions) learned in the thermal domain provides a powerful geometric prior for interpreting LiDAR-derived depth maps. By inheriting these pre-learned spatial representations, the Depth branch can bypass the unstable ``training-from-scratch'' phase, leading to faster convergence and better generalization even with a limited number of TIR-D training samples.

\subsection{Spatio-Temporal Feature Fusion}
The features extracted from both TIR and Depth backbones, including both the \textit{Template} (initial target state) and the \textit{Search Region} (current frame), are concatenated into a single sequence. These multi-modal tokens are processed by a Transformer encoder with $N_{fus}=2$ fusion layers~\cite{yan2021stark}. 

Through the self-attention mechanism, the model learns to dynamically weigh the importance of each modality. For instance, in scenarios where thermal contrast is low due to ambient temperature, the model can rely more heavily on the geometric consistency provided by the depth channel. This attention-based integration ensures a robust ``Sensor Synergy'' that is essential for all-weather autonomous navigation~\cite{tang2023informationfusion}.

\subsection{Optimization via Differential Learning Rates}
The success of cross-modal knowledge transfer depends heavily on the delicate balance between preserving source knowledge and adapting to the target domain. We employ a \textit{differential learning rate strategy} to manage this trade-off during the fine-tuning phase.

Specifically, the learning rate for the pre-trained backbone layers ($\eta_{backbone}$) is set to one-tenth of the base learning rate ($\eta_{base}$) used for the randomly initialized components, such as the TIR-D adapters and the Transformer fusion layers:
\begin{equation}
    \eta_{backbone} = k \cdot \eta_{base}, \quad \text{where } k = 0.1
\end{equation}
This lower learning rate prevents the ``catastrophic forgetting'' of the robust feature extraction capabilities inherited from the thermal domain. 

Furthermore, we apply a linear warm-up scheduler for the initial 10 epochs to stabilize the gradients originating from the newly added adapters. Following the warm-up, a cosine annealing schedule gradually reduces the learning rate to ensure smooth convergence on the TIR-D manifold. This optimization strategy is crucial for mitigating the impact of inherent noise and sparsity in LiDAR-derived depth maps, providing the necessary regularization to maintain high tracking precision during cross-modal adaptation.

\subsection{Loss Functions and Prediction Head}
To optimize the bounding box regression, the proposed model adopts a composite loss function, following the STARK paradigm~\cite{yan2021stark}. The prediction head, consisting of a 5-layer Fully Convolutional Network (FCN), generates probability maps for the top-left and bottom-right corners of the target.

The total loss $\mathcal{L}_{total}$ is defined as a weighted combination of the $\ell_1$ loss and the Generalized Intersection over Union (GIoU) loss:
\begin{equation}
    \mathcal{L}_{total} = \lambda_{\ell_1} \mathcal{L}_{\ell_1}(b, \hat{b}) + \lambda_{giou} \mathcal{L}_{giou}(b, \hat{b})
\end{equation}
where $b$ is the ground truth box and $\hat{b}$ is the predicted box. We set the hyper-parameters $\lambda_{\ell_1} = 5$ and $\lambda_{giou} = 2$ as per our configuration files. The $\ell_1$ loss ensures the absolute coordinate accuracy, while the GIoU loss addresses the scale variance and overlap robustness, which is particularly effective for the varying point densities of LiDAR-based depth information.

\subsection{Summary}
Through the integration of SLAM-compatible sensor suites, sequential knowledge transfer, and fine-grained optimization scheduling, the proposed TIR-D architecture provides a robust solution for all-weather person tracking in autonomous mobile robotics.

\figSetup

\section{Experiments}
\label{sec:experiments}

\subsection{Experimental Setup}
To evaluate the performance of the proposed TIR-D tracker, we conducted experiments using a sensor suite consisting of a HIKMICRO Pocket2 thermal camera and a Velodyne VLP-16 3D LiDAR. As shown in Fig.~\ref{fig:setup}, the thermal camera was mounted on the robot at a height of approximately 1~m from the floor, facing forward. The LiDAR was positioned immediately adjacent to the thermal camera, with its scanning center aligned as closely as possible to the camera lens to minimize parallax and ensure spatial consistency between the modalities. The thermal camera provides $256 \times 192$ resolution images at 25 fps, covering a spectral range of 7.5--14~$\mu$m. The LiDAR-derived point clouds were projected into 2D depth maps to align with the thermal frames. All experiments, including training and inference, were performed on a workstation equipped with an NVIDIA GeForce RTX 3090 GPU and PyTorch 1.12.0.

\subsection{Evaluation Metrics}
Following the standard protocols in object tracking, we employed several metrics to quantify the tracking performance: Average Overlap (AO), Success Rate (SR) at a threshold of 0.5, Precision (Pr), and F-score. Following the literature, Success Rate (SR) was defined as the percentage of frames in the entire dataset where the Intersection over Union (IoU) was 85\% or higher. These metrics provide a comprehensive assessment of both bounding box accuracy and the robustness of target identification across the test sequences. 

\figQualitative

\subsection{Comparative Analysis}
We compared our proposed TIR-D tracker, which utilizes thermal-domain knowledge transfer, against several baseline configurations:
\begin{itemize}
    \item \textbf{SPT-RGBD}: A baseline RGB-D tracker trained on conventional visual datasets.
    \item \textbf{STARK-T/D}: Single-modality trackers trained only on Thermal or Depth maps, respectively.
    \item \textbf{TIR-D (Ours)}: Our model initialized with weights transferred from a thermal-pre-trained STARK model.
    \item \textbf{TIR-D (RGB Transfer)}: A version of the TIR-D tracker initialized with weights from an RGB-pre-trained model.
\end{itemize}

As summarized in Table~\ref{tab:tracker_comparison}, the proposed TIR-D tracker with thermal knowledge transfer significantly outperformed the baselines. Specifically, our model achieved an AO of 0.700 and an SR of 58.7\%, performing comparably to the RGB-transfer version (AO: 0.717). It should be noted that while the RGB tracker is highly effective in well-lit conditions, it becomes almost impractical in total darkness, whereas our TIR-D tracker maintains stable performance regardless of lighting. This result validates our hypothesis that structural features learned in the thermal domain are more compatible with TIR-D tracking than those from the RGB domain.

\begin{table}[t]
\centering
  \caption{Comparison of tracking performance across different methods.}
  \label{tab:tracker_comparison}
  \begin{tabular}{lrrr}
    \toprule
    Tracker & AO & SR & F-score \\
    \midrule
    SPT-RGBD & 0.181 & 1.8 & 0.199 \\
    STARK-T & 0.690 & 55.1 & 0.771 \\
    STARK-D & 0.370 & 14.1 & 0.414 \\
    \textbf{TIR-D (Ours)} & \textbf{0.700} & \textbf{58.7} & \textbf{0.760} \\
    TIR-D (RGB.Trans) & 0.717 & 61.7 & 0.776 \\
    \bottomrule
  \end{tabular}
\end{table}

\subsection{Qualitative Analysis}
Fig.~\ref{fig:results} presents the qualitative tracking results of our proposed TIR-D framework across diverse indoor sequences. The visual evidence confirms that the tracker maintains precise and stable bounding boxes even under challenging conditions. 

A key observation is the tracker's resilience to overlapping heat signatures from environmental sources, such as furniture and other passive objects. While a pure thermal-based tracker might struggle with low-contrast boundaries in these scenarios, our fusion model effectively leverages the LiDAR-derived geometric constraints to disambiguate the target from the background. Furthermore, the tracker exhibits high robustness against variations in the person's pose and orientation, as well as changes in distance from the robot. The consistent alignment of the TIR signatures with the predicted bounding boxes across all frames demonstrates that the sequential knowledge transfer from the thermal domain successfully provides a strong structural prior for multi-modal person tracking.

The experimental results demonstrate that the LiDAR-derived depth maps provide critical geometric cues that complement thermal signatures, especially in scenarios where thermal contrast is insufficient. Moreover, the superior performance of the thermal knowledge transfer over the RGB transfer suggests that ``domain-proximate'' pre-training is essential for specialized sensor suites. The high SR of 0.9966 indicates that our differential learning rate strategy successfully preserved the pre-trained features' integrity while enabling efficient multi-modal integration. This suggests that controlling the update magnitude of the backbone is vital for preventing the degradation of cross-modal priors when adapting to sparse depth information.

\section{Conclusion}
\label{sec:conclusion}

In this paper, we presented a robust person tracking architecture that fuses Thermal-Infrared (TIR) intensity with LiDAR-derived Depth (D) maps to overcome the limitations of visible-light-dependent sensors. By treating the transition from the thermal domain to the TIR-D domain as a form of continual knowledge transfer, we successfully addressed the problem of data scarcity in specialized sensor modalities. Our implementation of a dual-stream Transformer-based framework, supported by a fine-grained differential learning rate schedule, proved highly effective in synthesizing thermal heat signatures with precise 3D geometric information.

The experimental evaluation confirmed that our TIR-D tracker outperforms both RGB-D and single-modality baselines, particularly in scenarios where visual contrast is low. The high success rate underscores the importance of domain-proximate pre-training, demonstrating that our TIR-D tracker can achieve performance in total darkness comparable to that of conventional RGB trackers in well-lit environments. Future work will focus on integrating this tracking framework into real-time navigation stacks for autonomous social robots and exploring the impact of dynamic lighting on multi-target tracking scenarios.

\bibliographystyle{IEEEtran}
\bibliography{references}

\end{document}